\documentclass[letterpaper, 10 pt, conference]{ieeeconf}  % Comment this line out if you need a4paper

\IEEEoverridecommandlockouts                              % This command is only needed if 
\usepackage{xcolor}

\usepackage{graphicx}

\usepackage{gensymb}
\usepackage[hyphens]{url}
\usepackage{multirow}
\usepackage{multicol}
\usepackage{tabularx}
\usepackage{dsfont}
\usepackage{subcaption}
\usepackage{caption}
\usepackage{booktabs}
\usepackage{dcolumn}
\usepackage{amssymb}
\usepackage{amsmath}
\usepackage{colortbl}
\usepackage{tikz}
\colorlet{lightgray}{gray!20}
\usepackage{physics}
\usepackage{graphicx}
\usepackage{float}
\usepackage{subfloat}
\usepackage{longtable}
\usepackage{threeparttable}
\usepackage{diagbox}
\usepackage{cite}
\usepackage{xspace}
\usepackage{pifont} % \ding symbols
\usepackage{booktabs}
\usepackage{multirow}

 \renewcommand{\paragraph}[1]{
    \vspace{2mm}
     \noindent\textbf{#1} 
 }

\definecolor{darkgreen}{rgb}{0,0.694,0.298}
\definecolor{purple}{rgb}{0.4,0.176,0.569}
\definecolor{royalblue}{RGB}{65,105,225}
\definecolor{americanrose}{rgb}{1.0, 0.01, 0.24}
\definecolor{applegreen}{rgb}{0.55, 0.71, 0.0}

\definecolor{ao(english)}{rgb}{0.0, 0.5, 0.0}
\definecolor{amaranth}{rgb}{0.9, 0.17, 0.31}

\usepackage{hyperref}
\hypersetup{
    colorlinks=true,    
    linkcolor=blue,
    citecolor=royalblue,
    filecolor=magenta,      
    urlcolor=cyan,
    pdftitle={Overleaf Example},
    pdfpagemode=FullScreen,
    }
    
\makeatletter
\DeclareRobustCommand\onedot{\futurelet\@let@token\@onedot}
\def\@onedot{\ifx\@let@token.\else.\null\fi\xspace}
 
\def\ie{\emph{i.e}\onedot}

\makeatother

\title{\LARGE \bf AdvGPS: Adversarial GPS for Multi-Agent Perception Attack}  

\author{Jinlong Li$^{1}$, Baolu Li$^{1}$, Xinyu Liu$^{1}$, Jianwu Fang$^{2}$, Felix Juefei-Xu$^{3}$, Qing Guo$^{4*}$, Hongkai Yu$^{1*}$  
\thanks{$^{1}$Cleveland State University.   
$^{2}$Xi'an Jiaotong University. 
$^{3}$New York University.
$^{4}$CFAR and IHPC, Agency for Science, Technology and Research (A*STAR), Singapore. This research is supported by NSF 2215388, and the National Research Foundation, Singapore, and DSO National Laboratories under the AI Singapore Programme (No: AISG2-GC-2023-008).
*Co-corresponding authors: tsingqguo@ieee.org and h.yu19@csuohio.edu.}
}

\begin{document}

\maketitle
\thispagestyle{empty}
\pagestyle{empty}

\begin{abstract}
The multi-agent perception system collects visual data from sensors located on various agents and leverages their relative poses determined by GPS signals to effectively fuse information, mitigating the limitations of single-agent sensing, such as occlusion.
However, the precision of GPS signals can be influenced by a range of factors, including wireless transmission and obstructions like buildings. Given the pivotal role of GPS signals in perception fusion and the potential for various interference, it becomes imperative to investigate whether specific GPS signals can easily mislead the multi-agent perception system.
To address this concern, we frame the task as an adversarial attack challenge and introduce \textsc{AdvGPS}, a method capable of generating adversarial GPS signals which are also stealthy for individual agents within the system, significantly reducing object detection accuracy.
To enhance the success rates of these attacks in a black-box scenario, we introduce three types of statistically sensitive natural discrepancies: appearance-based discrepancy, distribution-based discrepancy, and task-aware discrepancy.
Our extensive experiments on the OPV2V dataset demonstrate that these attacks substantially undermine the performance of state-of-the-art methods, showcasing remarkable transferability across different point cloud based 3D detection systems.
This alarming revelation underscores the pressing need to address security implications within multi-agent perception systems, thereby underscoring a critical area of research.
The code is available at https://github.com/jinlong17/AdvGPS.
\end{abstract}

\section{Introduction}\label{Sec:intro}
Although the single-agent perception system gets advanced performance in many autonomous driving scenarios, it still has considerable sensing limitations due to the challenges of occlusion and perception range. Benefited from the recent research of multi-agent perception system, the visual data from sensors on nearby agents can be shared to the ego-agent as information fusion to improve the perception range and overcome occlusion challenges~\cite{xu2022opv2v}. During the visual data sharing from nearby agents to the ego-agent, one important and necessary step is to project the visual data from the coordinate systems of nearby agents to the uniform coordinate system of the ego-agent via homogeneous transformation~\cite{xu2023v2v4real}. This homogeneous transformation based uniform coordinate projection requires their  relative pose (localization \& heading) from nearby agents to ego agent~\cite{xu2021opencda,xu2023v2v4real,li2023learning}, which is  determined by their GPS signals.    

\begin{figure}[]
    \begin{centering}        \includegraphics[width=1\columnwidth]{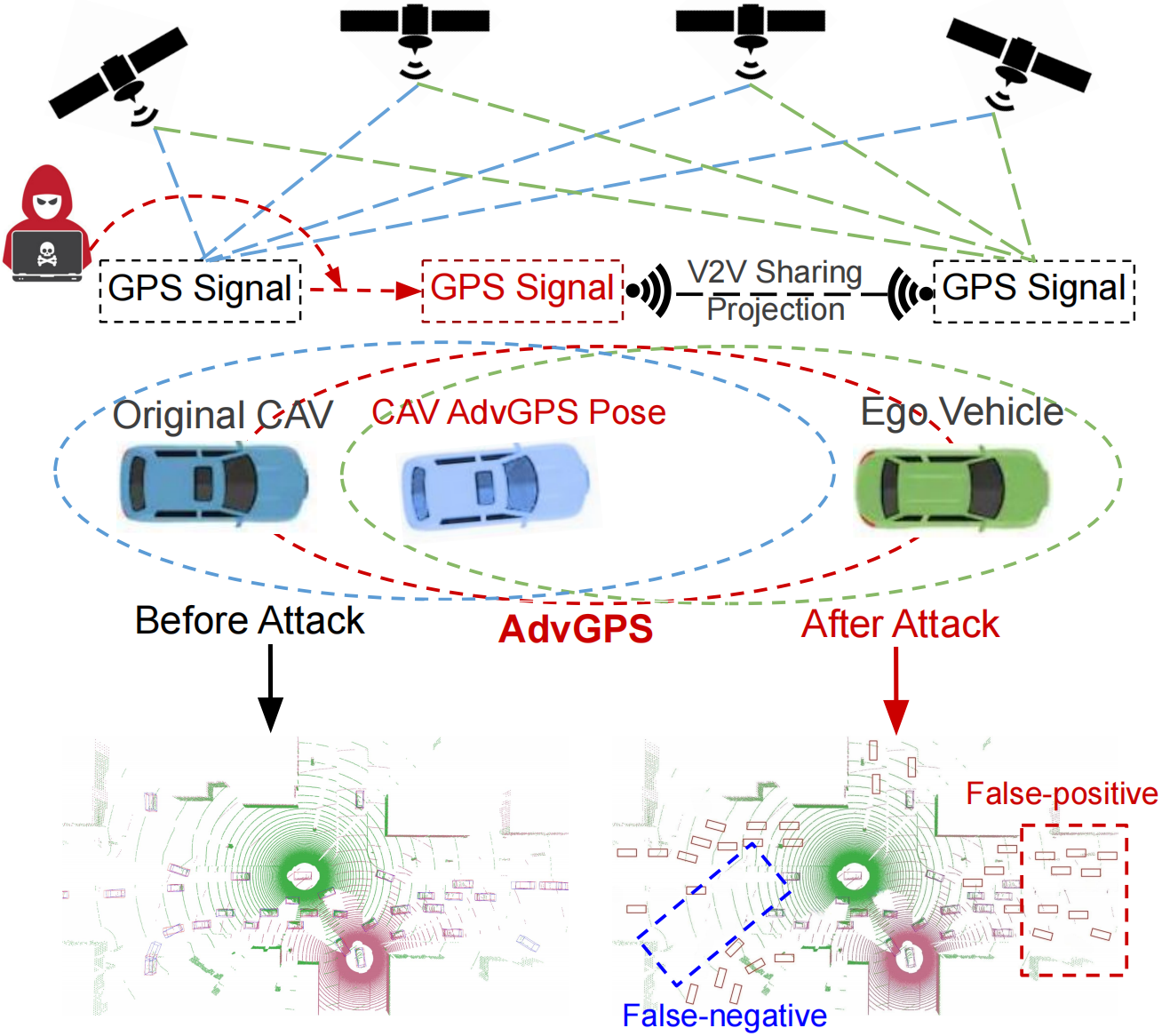}
        \par\end{centering}
    \caption{\textbf{Illustration of AdvGPS for multi-agent perception attack}. Here we use Vehicle-to-Vehicle (V2V) cooperative perception in autonomous driving as an example. Ego vehicle might receive the shared visual 
    information from other CAVs with the adversarial GPS signal, leading to significant false-negative and false-positive detection errors.}
    % \caption{\textbf{Illustration of AdvGPS for multi-agent perception attack}. Here we use V2V cooperative perception in autonomous driving as an example. Ego vehicle receives the shared visual information from other CAVs with adversarial GPS signal, which causes an adversarial shift of the projected point cloud leading to performance drops significantly.}
    \label{fig:attack}
\end{figure}
 
However, it is a common sense that the GPS signals might have unavoidable errors in the real world~\cite{yang2023anomaly,xu2023v2v4real}. 
GPS signal accuracy can be affected by a variety of factors, such as  wireless transmission, building obstacles and so on. This paper uses Vehicle-to-Vehicle (V2V) cooperative perception in autonomous driving as a study example of the multi-agent perception system. The previous work~\cite{yang2023anomaly} shows that simply adding random GPS signal noises in a specific range could result in spoofing attacks to Connected and Automated Vehicles (CAV). In the research domain of V2V cooperative perception, whether adversarial GPS signals can easily mislead the multi-agent perception system is never studied before as an open question.

To investigate the above open question, we model the task as an adversarial attack challenge and introduce a novel method \textbf{AdvGPS} capable of generating adversarial GPS signals which are also stealthy for nearby CAVs, significantly reducing object detection accuracy of the ego vehicle. As illustrated in Fig.~\ref{fig:attack}, when the adversarial GPS signals within a small/stealthy range are added to the nearby CAVs, it will  generate significant false-negative (missing) and false-positive detection errors for the ego vehicle. We formulate the AdvGPS attack in the black-box setting, because the prior knowledge of the target perception model is usually unknown. To enhance the success rates of the AdvGPS attack in the black-box scenario, we introduce three types of statistically sensitive natural discrepancies for multi-agent perception attack: appearance-based discrepancy, distribution-based discrepancy, and task-aware discrepancy.

Using the publicized dataset OPV2V~\cite{xu2022opv2v}, we conducted extensive experiments and the experimental results demonstrate that the proposed AdvGPS attacks substantially undermine the performance of state-of-the-art V2V cooperative perception methods, showcasing remarkable transferability across different point cloud based 3D detection systems. This alarming discovery emphasizes the critical need to confront security implications in multi-agent perception systems, thus highlighting a pivotal research area. Our contributions of this paper can be summarized as follows.

\begin{itemize}
    \item To the best of our knowledge, we propose the  \textbf{first research} of adversarial GPS signals which are also stealthy for the V2V cooperative perception attacks, denoted as AdvGPS.

    \item We propose three statistically sensitive natural discrepancies in AdvGPS to enhance the multi-agent perception attack in the black-box scenarios, \textit{i.e.}, appearance-based discrepancy, distribution-based discrepancy, and task-aware discrepancy.

    \item The experimental results on the publicized OPV2V dataset demonstrate that our AdvGPS attacks substantially undermine the performance of state-of-the-art methods and  
    show outstanding transferability across different point cloud based 3D detection systems.  
\end{itemize}

\section{Related Work}\label{Sec:Related_Work}

\noindent \textbf{Cooperative Perception in V2V.}
% Multi-vehicle perception systems aim to surpass single-vehicle limitations through multi-vehicle information utilization. Researchers commonly develop collaboration modules to enhance efficiency and performance. Typically, three schemes employ basic modules for multi-vehicle observation aggregation: raw data fusion at the input stage (early fusion), feature fusion during processing (intermediate fusion), and output fusion (late fusion).
% State-of-the-art methods usually share the intermediate neural features with contextual knowledge of the environment, as they can achieve the best trade-off between accuracy and bandwidth requirement~\cite{xu2022opv2v,xu2022v2x}. 
% Attfuse~\cite{xu2022opv2v} and V2VAM~\cite{} utilize the attention mechanism to fuse the multi-vehicle features for 3D object detection, and the popular vision Transformer architecture are also introduced to capturing complex spatial interactions among multiple agents, \textit{e.g.} V2X-ViT~\cite{xu2022v2x}, CoBEVT~\cite{xu2022cobevt}, ~\cite{},  
% Although these methods have demonstrated impressive performance in V2V perception, the adversarial robustness of them is underexplored with GPS attack.
 Multi-vehicle perception systems have arisen to address the inherent limitations of single-vehicle sensing by harnessing information exchange among multiple vehicles. Researchers have frequently incorporated collaboration modules to enhance efficiency and overall system performance. Typically, these systems employ three fundamental schemes for multi-vehicle observation aggregation: raw data fusion at the input stage, intermediate feature fusion during processing, and output fusion.
State-of-the-art approaches often opt for sharing intermediate neural features augmented with contextual information about the environment. This choice makes  a favorable balance between accuracy and bandwidth requirements~\cite{xu2022opv2v,xu2022v2x}. Notably, methods such as Attfuse~\cite{xu2022opv2v} and V2VAM~\cite{10077757} leverage attention mechanisms to fuse multi-vehicle features for 3D object detection. Additionally, the integration of the popular Vision Transformer has been introduced to capture complex spatial interactions among multiple agents, like V2X-ViT~\cite{xu2022v2x}, and CoBEVT~\cite{xu2022cobevt}.
While these methods have exhibited impressive performance in the context of V2V perception, their robustness against adversarial GPS attacks remains an under-explored research. \\
\noindent \textbf{Deployment of Cooperative Perception.}
% Multi-Agent perception systems offer advantages but introduce challenges like localization errors,  communication latency, adversarial attack, and lossy communication, which can undermine collaboration benefits~\cite{xu2022bridging,xu2022v2x}.
% To enhance robustness, several approaches have been proposed. 
% V2X-ViT~\cite{xu2022v2x} utilizes a ViT to withstand GPS localization errors and sensing information delays in intermediate collaboration. To address localization errors, \cite{vadivelu2021learning} proposed a pose regression module that learns a correction parameter to predict the true relative transformation from noisy data. The LC-aware Repair Network (LCRN)~\cite{li2023learning}  is introduced to enhance collaborative perception robustness under lossy communication conditions to address packet loss problem in communication. a model-agnostic framework~\cite{xu2023model} is proposed to address model heterogeneity in collaborative perception. 
% However, the cooperative perception performance relies on the GPS signal, which is vulnerable to attacks
% in the real world. This challenge consequently results in a substantial degradation of cooperative perception performance. In this paper, we investigate the GPS attack for 3D object detection task on V2V cooperative perception.
Multi-agent perception systems offer numerous advantages but are accompanied by challenges such as localization errors, communication latency, adversarial attacks, and lossy communication. These development challenge can easily diminish the benefits of collaborations~\cite{xu2022bridging,xu2022v2x}. Several strategies have been proposed to enhance robustness.
For instance, V2X-ViT employs a Vision Transformer to mitigate GPS localization errors and sensing information delays during intermediate collaboration~\cite{xu2022v2x}. To address localization errors, \cite{vadivelu2021learning} proposes a pose regression module that learns correction parameters to predict accurate relative transformations from noisy data. The Lossy Communication-aware Repair Network (LCRN)~\cite{li2023learning} is introduced to address packet loss issues in communication. A Multi-agent Perception Domain Adaption (MPDA) framework~\cite{xu2023bridging} is proposed for cooperative perception to bridge the domain gap for multi-agent perception. Nevertheless, V2V cooperative perception performance relies heavily on GPS signals, which might be susceptible to real-world attacks. This challenge might lead to a significant degradation in perception performance. This paper investigates GPS attacks in the context of 3D object detection tasks within V2V cooperative perception. \\
% \noindent \textbf{Adversarial Attack in Perception.}
% \noindent \textbf{GPS Attack in Vehicle.}  Adversarial attack
% A model-agnostic framework~\cite{xu2023model} has also been proposed to handle model heterogeneity in collaborative perception. 
\noindent \textbf{Adversarial Attack.} 
% Current adversarial attack have received the more attention, which aims to generate adversarial perturbations and use them to fool the target model into predicting an incorrect label. Adversarial attack can be classified into white-box attack and black-box attacks. White-box attacker has full information about the deep learning models while black-box attacker is usually less effective than white box attack since attackers have no access to the target models. In autonomous vehicle, GPS spoofing is a significant threat to vehicle localization, which involves broadcasting false signals to mislead a vehicle.  Previous research has demonstrated the GPS spoofing can endanger the autonomous vehicles driving in the real world~\cite{}. And the constant bias and gradual drift attacks is mostly used GPS attack methods~\cite{}, ~\cite{xu2023sok} introduced a position altering attack (PAA) which uses a GPS spoofer to mislead the Vehicle to a fake position.
% In this paper, we introduce the GPS original physical attack to cooperative perception system, and study if current coperative perception methods are robust enough under GPS attack.
Adversarial attacks have garnered significant attention, primarily focused on generating perturbations designed to mislead deep learning models into producing incorrect predictions \cite{hou2023evading}. These attacks can be categorized into two main types: white-box attacks \cite{carlini2017towards,li2021fooling}, where the attacker possesses full knowledge of the target model, and black-box attacks \cite{cheng2019improving,shi2019curls}, which are generally less effective as the attacker lacks access to the target model's internal details. In autonomous vehicles, GPS spoofing poses a substantial threat to vehicle localization, involving the transmission of false signals to deceive a vehicle's GPS system. Previous research has highlighted the dangers of GPS spoofing for autonomous vehicles operating in real-world scenarios \cite{li2021fooling,xu2023sok}. Notably, constant bias and gradual drift attacks are common GPS attack methods \cite{yang2023anomaly}. Additionally, recent work \cite{xu2023sok} introduced a position-altering attack (PAA), employing a GPS spoofer to mislead a vehicle to a fictitious location. In this paper, we introduce the new concept of GPS attacks into the V2V cooperative perception systems.

\section{V2V Perception Pipeline and Motivation}
\label{sec: preliminary}

\begin{figure}[]
    \begin{centering}
        \includegraphics[width=1\columnwidth]{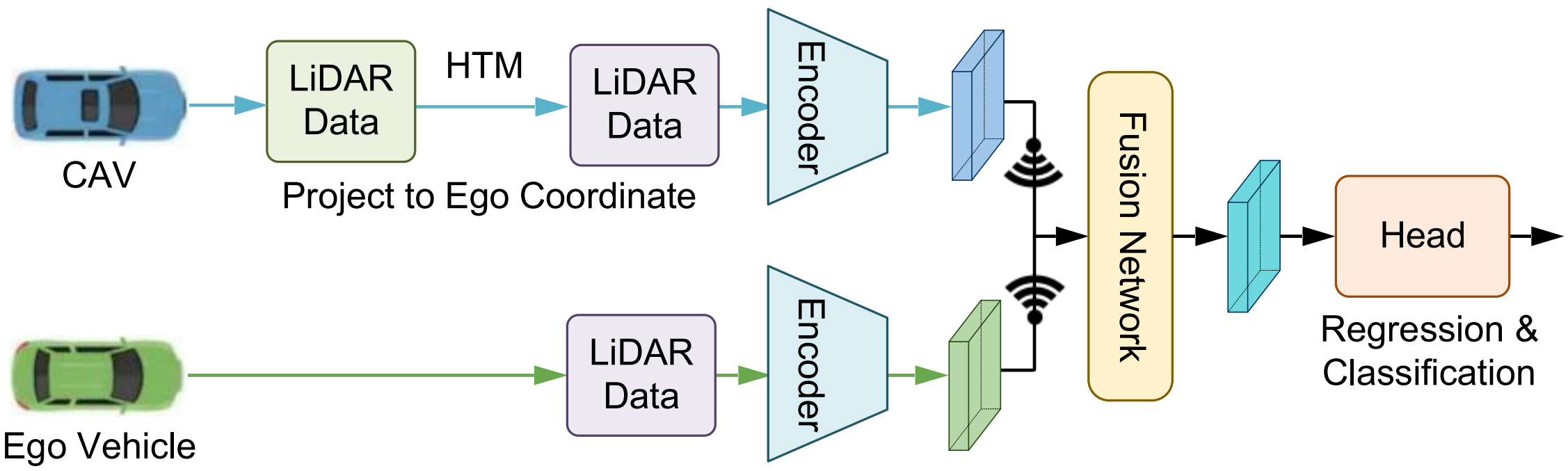}
        \par\end{centering}
    \caption{\textbf{Illustration of V2V cooperative perception pipeline with LiDAR data coordinate  projection from CAV to Ego.}}
    \label{fig:V2V_pipeline}
\end{figure}

\begin{figure*}[htbp]
    \begin{centering}
        \includegraphics[width=0.95\textwidth]{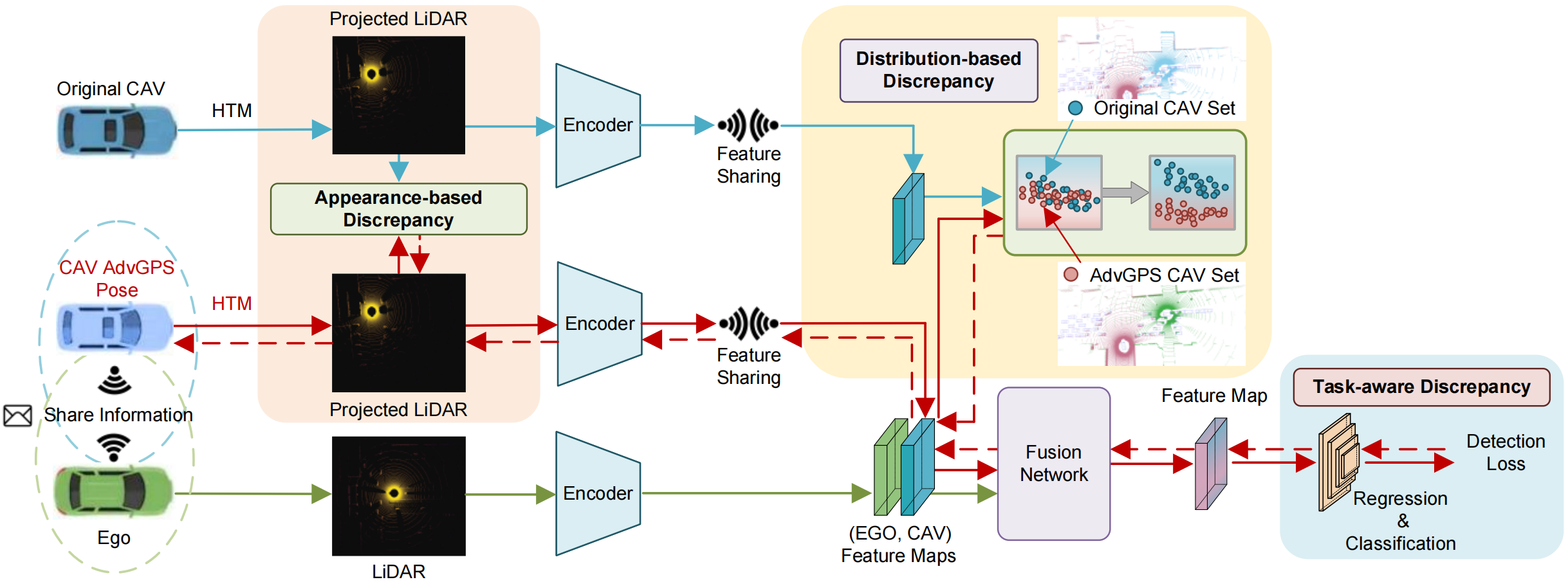}
        \par\end{centering}
    \caption{\textbf{Pipeline of AdvGPS for multi-agent cooperative perception attack}. Dash lines indicate the back propagation.}
    \label{fig:AdvGPS}
\end{figure*}

The V2V perception pipeline, as illustrated in Fig.~\ref{fig:V2V_pipeline}, initiates by selecting an ego vehicle from among the CAVs to create a spatial graph that encompasses nearby CAVs within the communication range. 
These nearby CAVs project their LiDAR data into the ego vehicle's coordinate frame using a Homogeneous Transformation Matrix (HTM) based on both the ego vehicle's and their own GPS poses.
Subsequently, the pipeline proceeds with feature extraction, where each CAV employs its own LiDAR feature extraction module. These extracted features are aggregated and fused via a feature fusion neural network. Finally, the fused feature maps are used for 3D bounding-box regression and classification, ultimately facilitating advanced cooperative perception in autonomous driving.

To elucidate this process with an example, let us define the ego vehicle's GPS pose as $G_{ego}$ and the GPS pose of a neighboring CAV as $G_{cav}$, where both poses contain six variables $[x, y, z, \theta_{x}, \theta_{y}, \theta_{z}]$. The CAV can perceive the environment via its own sensor and get the point cloud data denoted as $\mathbf{P}_{cav} \in \mathbb{R}^{4 \times m}$ that contains a set of 3D points $\{ P_{i} \mid i=1,2,\cdots,m \}$, where each point $P_{i}$ is a vector and contains the homogeneous coordinates $[x, y, z, 1]$.
We can use a Homogeneous Transformation Matrix (HTM) $\mathbf{T}_{cav \rightarrow ego} \in \mathbb{R}^{4 \times 4}$ to project/transform the point cloud of CAV to the uniform coordinate system of ego vehicle by 
\begin{equation}
    \mathbf{P}_{cav \rightarrow ego} =  \mathbf{T}_{cav \rightarrow ego}  \mathbf{P}_{cav},  
    \label{eq:cav_2_ego}
\end{equation}
where $\mathbf{P}_{cav \rightarrow ego}$ denotes the transformed point cloud of $\mathbf{P}_{cav}$ in the ego coordinate system. The HTM $\mathbf{T}_{cav \rightarrow ego}$ format is $\begin{bmatrix}
    \mathbf{r} & \mathbf{t} \\
    \mathbf{0}_{1 \times 3} & 1
\end{bmatrix}$, where $\mathbf{r} \in \mathbb{R}^{3 \times 3}$ and $\mathbf{t} \in \mathbb{R}^{3 \times 1}$ are the rotation matrix and translation matrix, respectively. Then, the HTM $\mathbf{T}_{cav \rightarrow ego}$ can be extracted/calculated via 
\begin{equation}
    \mathbf{T}_{cav \rightarrow ego} 
    = \text{HTM$_{\text{Extract}}$}(G_{ego}, G_{cav}), 
    \label{eq:HTM}
\end{equation}
%
% Note that the HTM represents the relative pose between the ego vehicle and CAV and is calculated by the two GPS signals.
where $\text{HTM}_{\text{Extract}}(\cdot)$ is an inverse function (matrix computation as defined in~\cite{xu2023v2v4real}) of homogeneous transformation to calculate the HTM based on the GPS poses of two entities, \ie, $G_{ego}$ and $G_{cav}$. 
According to the pipeline, GPS signals play a critical role in the V2V perception.
Nevertheless, the accuracy of GPS signals is easily affected by diverse factors, and we aim to explore whether some specific GPS patterns could fool the V2V perception easily.
To this end, we formulate the task from the view of an adversarial attack and propose the AdvGPS against the multi-agent perception in Sec.~\ref{Sec:Method}. 
Our goal is to search the adversarial GPS of CAV (\ie, $\widehat{G}_{cav} \in [\hat{x}, \hat{y}, \hat{z}, \hat{\theta}_{x}, \hat{\theta}_{y}, \hat{\theta}_{z}]$) that can lead to an adversarial HTM via Eq.~\ref{eq:HTM}. 
In the real world, adversarial GPS signals can be concealed within the range of actual stealthy GPS errors.

\section{AdvGPS against Multi-agent Perception}
\label{Sec:Method} 

\subsection{Overview}
\label{subsec:overview}

We propose AdvGPS and present the whole pipeline in Fig.~\ref{fig:AdvGPS}. 
Specifically, we formulate V2V cooperative perception system $\phi(\cdot)$ for LiDAR-based 3D object detection as
\begin{align}
    \phi(\mathbf{P}_{cav},\mathbf{P}_{ego},G_{cav}) =  \varphi(\text{Fusion}(\mathbf{F}_{cav \rightarrow ego}, \mathbf{F}_{ego})),
    \label{eq:v2v_total}
\end{align}
where $\mathbf{F}_{cav \rightarrow ego}=\text{Encoder}(\mathbf{T}_{cav \rightarrow ego}  \mathbf{P}_{cav})$ and $\mathbf{F}_{ego}=\text{Encoder}(\mathbf{P}_{ego})$ denote the features of the two point clouds $\mathbf{P}_{cav \rightarrow ego}$ and $\mathbf{P}_{ego}$, $\text{Fusion}(\cdot)$ is a network to fuse the two features, and $\varphi(\cdot)$ is the regression and detection header for object detection~\cite{xu2022opv2v}. 
% and $\phi(\cdot)$ is the head for object detection.
%
It aims to exploit adversarial attacks on GPS signals to deceive any point cloud based 3D detection models in a black-box setting. We calculate an imperceptible noise-like perturbation under the guidance of a classic 3D object detection model and add it to the original CAV GPS pose to obtain an AdvGPS pose. This corrupted AdvGPS pose can then mislead any other detection models in a black-box setting. 
% Unlike traditional adversarial attacks based on additive perturbations, our adversarial GPS attack is non-additive and enjoys better transferability and stealthiness. 
Our AdvGPS attack perturbation is stealthy to be confused as normal GPS signal errors. 
To enhance the success rates of attacks in black-box scenarios, we consider three objectives to optimize adversarial GPS: appearance-based discrepancy, distribution-based discrepancy, and detection task-aware discrepancy. 
We define the optimization over the adversarial GPS signal $\widehat{G}_{cav}$ as:
\begin{equation}
    \mathrm{arg} \max\limits_{\widehat{G}_{cav}} ( \lambda D_{app} + \omega D_{dist} + \xi D_{task}) 
    \label{eq:AdvGPS}
\end{equation}
% \begin{equation}
%         \mathrm{subject\ to}\ \Vert (\hat{x}, \hat{y}) -(x,y) \Vert < \epsilon_{(x, y)},
%     \label{eq:xy}
% \end{equation}
% \begin{equation}
%          \Vert (\hat{z}) - (z) \Vert < \epsilon_{(z)},\\
%         \Vert (\hat{\theta}_{x}, \hat{\theta}_{y}, \hat{\theta}_{z}) - (\theta_{x}, \theta_{y}, \theta_{z})  \Vert < \epsilon_{(\theta_{x}, \theta_{y}, \theta_{z})},
%     \label{eq:z}
% \end{equation}

\begin{equation}
    \begin{aligned}
        \text{subject to} \quad & \| [\hat{x}, \hat{y}] - [x,y] \|_{\infty} < \epsilon_{x, y} \\
                               & \|\hat{z} - z\|_{\infty} < \epsilon_{z} \\
                               & \| [\hat{\theta}_{x},  \hat{\theta}_{y},\hat{\theta}_{z}] - [\theta_{x}, \theta_{y},
                               \theta_{z}]\|_{\infty} < \epsilon_{\theta_{x}, \theta_{y}, \theta_{z}},
    \end{aligned}
    \label{eq:constraints}
\end{equation}
where $\lambda$, $\omega$, and $\xi$ are the balance weights and 1s are used for them in our experiments. As the three responses of the V2V cooperative perception system $\phi(\cdot)$ defined in Eq.~\ref{eq:v2v_total} using $\widehat{G}_{cav}$,  $D_{app}$ represents the appearance-based discrepancy before and after the GPS attack, $D_{dist}$ is used to measure the distribution-based discrepancy between original CAV set and AdvGPS CAV set, and $D_{task}$ accounts for the task-aware discrepancy using a classic 3D object detection model. Our goal is to maximize the objective function by tuning AdvGPS pose $\widehat{G}_{cav}$, which includes six parameters $[\hat{x}, \hat{y}, \hat{z}, \hat{\theta}_{x}, \hat{\theta}_{y}, \hat{\theta}_{z}]$, meanwhile we constrain the GPS signal perturbation within the range of normal  stealthy GPS errors in real world via the $\epsilon$ values by Eq.~\ref{eq:constraints}.

\subsection{Appearance-based Discrepancy}
\label{subsec:appdist}

The appearance-based discrepancy focuses on the differences between the original CAV projected point cloud $\mathbf{P}_{cav \rightarrow ego}^{ori}$ and the AdvGPS-based projected point cloud $\mathbf{P}_{cav \rightarrow ego}^{adv}$. The cooperative perception performance is directly affected by the appearance difference when fusing into the ego vehicle's perception system. To efficiently disrupt cooperative perception performance through appearance difference, our $D_{app}$ can be defined as the appearance difference between these two point clouds by tuning the AdvGPS pose $\widehat{G}_{cav}$ via the mathematical expression:
~
% \begin{equation}
%     \mathrm{arg} \max\limits_{_{adv}p_{cav}}\ D_{appe} (\mathbf{P}_{cav \rightarrow ego}^{real},\mathbf{P}_{cav \rightarrow ego}^{adv}).
%     \label{eq:Appe}
% \end{equation}
\begin{equation}
    D_{app} = L_{MSE} (\mathbf{P}_{cav \rightarrow ego}^{ori},\mathbf{P}_{cav \rightarrow ego}^{adv}),
    \label{eq:Appe}
\end{equation}
where $L_{MSE} (\cdot)$ is the mean squared error (MSE) loss between the two point clouds.
\subsection{Distribution-based Discrepancy}
\label{subsec:disdist}

After encoding, we obtain a set of original features by  the GPS poses of Original CAV Set and a set of attacked features by the adversarial GPS poses of AdvGPS CAV Set. We investigate the correlations between statistical differences and the distribution of these features. Some previous studies~\cite{hou2023evading} have found that statistical differences are positively associated with distribution differences. To maximize the discrepancy distance between the feature distributions of the original features $\mathbf{F}^{ori}$ and attacked features $\mathbf{F}^{adv}$, we use maximum mean discrepancy (MMD)~\cite{JMLR:v13:gretton12a}  to measure the distance of them. Let $\mathcal{F}^{ori} = \{ \mathbf{F}_{i}^{ori}\}$ and $\mathcal{F}^{adv}=\{ \mathbf{F}_{i}^{adv}\}$ represent a set of original and adversarial features after encoding. 
Our goal is to generate adversarial GPS poses capable of degrading cooperative detection performance in terms of feature distributions. This can be defined as:
~
\begin{equation}
     D_{dist} = L_{MMD} (\mathcal{F}^{ori}, \mathcal{F}^{adv}),
    \label{eq:Dist}
\end{equation}
where $L_{MMD} (\cdot)$ represents the MMD loss between the two intermediate features after encoding.

\subsection{Task-aware Discrepancy}
\label{subsec:taskdis}

Intuitively, given the LiDAR point clouds from ego vehicle and surrounding CAVs (\textit{i.e.}, $\mathbf{P}_{ego}$ and $\mathbf{P}_{cav})$, a pre-trained cooperative detection model is used to simulate the  target to be attacked. In our experiments, the pre-trained cooperative detection model uses the classic 3D object detection model VoxelNet~\cite{zhou2018voxelnet} as encoder and a simple self-attention module as the fusion network. Then, based on the Eq.~\ref{eq:v2v_total}, we get the predicted detection results and calculate the loss according to the ground truth (\ie, $Y$). $L_{DET}$ is the task-aware discrepancy to mislead the pre-trained cooperative detection model, which can be formulated as 
% \begin{equation}
%     D_{task} = L{detection} (M(\mathbf{T}_{cav \to ego}(\mathbf{P}_{cav}), \mathbf{P}_{ego}), Y).
%     \label{eq:}
% \end{equation}
\begin{equation}
    D_{task} = L_{DET} (\phi(\mathbf{P}_{cav},\mathbf{P}_{ego},\widehat{G}_{cav}), Y),
    \label{eq:task}
\end{equation}
where $L_{DET} (\cdot)$ is denoted as the detection loss including regression and classification of 3D bounding boxes. Please note that the above pre-trained cooperative detection model to simulate target to be attacked is classic and simple in 3D point cloud detection,  which is different with other state-of-the-art V2V cooperative perception models, so our AdvGPS   still follows the black-box attack setting.

\subsection{Implementation Details}
\label{subsec:impl}

To implement our AdvGPS for V2V cooperative perception attack, we follow the general adversarial attack procedure:
\ding{182} We set the additive perturbation  $\mathbf{g}$ to $\widehat{G}_{cav}$ and obtain the original CAV pose to calculate the HTM. 
\ding{183} Based on the HTM, we calculate $D_{app}$ for the projected point cloud.
\ding{184} We feed the projected point cloud into an encoder to generate the intermediate feature and calculate $D_{dist}$.
\ding{185} The intermediate features are then inputted  into the fusion network to calculate $D_{task}$.
% 5) We perform back-propagation and compute the gradients of $_{adv}p_{cav}$ with respect to the loss function.
\ding{186} We perform back propagation and compute the gradients of $\mathbf{g}$ with respect to the loss functions.
% 6) The sign of the gradients is used to update $_{adv}p_{cav}$ by multiplying them with a step size.
\ding{187} The sign of the gradients is used to update additive perturbation $\mathbf{g}$   by multiplying them with a step size, then adding $\mathbf{g}$  to the $\widehat{G}_{cav}$.
\ding{188} We calculate a new synthesized $\mathbf{T}_{cav \rightarrow ego}$ via Eq.~\ref{eq:HTM} and repeat steps \ding{183} to \ding{187} for a defined number of iterations.
% For loss functions, $D_{Appe}$, $D_{Dist}$, and $D_{Task}$ can be implemented using mean squared error (MSE) loss, maximum mean discrepancy (MMD) loss, and detection loss (classification and regression), respectively.
We set the number of iterations to 10.
% and use the $\infty$ norm for $\epsilon_{(x, y)}$, $\epsilon_{(z)}$, and $\epsilon_{(\theta_{x}, \theta_{y}, \theta_{z})}$.
To conceal within the range of actual GPS errors, we follow the statistics presented in~\cite{chiang2020performance}, 
%
% which indicate that the real mean/standard deviation of position error and heading error are $1.118/3.686$ meters and $0.141/0.529$ degrees, respectively. We set $\epsilon_{(x, y, z)}$ and $\epsilon_{(\theta_{x}, \theta_{y}, \theta_{z})}$ as 1.118 and 0.141, respectively.
%
which indicates that the real-world GPS signal's average localization  error, height error, and heading error are $1.118$ meters,  $1.395$ meters and $0.141$ degrees, so we set $\epsilon_{x, y}$, $\epsilon_{z}$, and $\epsilon_{\theta_{x}, \theta_{y}, \theta_{z}}$ as 1.118, 1.395, and 0.141 respectively to avoid significant offsets.

\section{Experiments}\label{Sec:Experiment}

% \usepackage{booktabs}
% \usepackage{multirow}
% \usepackage{graphicx}
\begin{table*}[]
\centering
\tiny
\caption{ \textbf{Quantitative results of GPS attack}: 3D detection performance on V2V Culver City  testing set of OPV2V dataset. We show the Average Precision (AP) at IoU=0.5. The best and second best attacked performance among five state-of-the-art cooperative perception methods are respectively highlighted using {\color{red}red} and {\color{blue}blue} color. }
\label{tab:gps_result}
\resizebox{0.95\textwidth}{!}{%
\begin{tabular}{@{}
>{\columncolor[HTML]{FFFFFF}}c 
>{\columncolor[HTML]{FFFFFF}}c 
>{\columncolor[HTML]{FFFFFF}}c 
>{\columncolor[HTML]{FFFFFF}}c 
>{\columncolor[HTML]{FFFFFF}}c 
>{\columncolor[HTML]{FFFFFF}}c 
>{\columncolor[HTML]{FFFFFF}}c 
>{\columncolor[HTML]{FFFFFF}}c 
>{\columncolor[HTML]{FFFFFF}}c 
>{\columncolor[HTML]{FFFFFF}}c 
>{\columncolor[HTML]{FFFFFF}}c 
>{\columncolor[HTML]{FFFFFF}}c 
>{\columncolor[HTML]{FFFFFF}}c @{}}
\toprule
   Model  &
   CAVs' Pose&
  No Attack &
  RBA~\cite{yang2023anomaly} &
  \multicolumn{2}{c}{\cellcolor[HTML]{FFFFFF}FSGM~\cite{43405}} &
  \multicolumn{2}{c}{\cellcolor[HTML]{FFFFFF}IFSGM~\cite{kurakin2018adversarial}} &
  \multicolumn{2}{c}{\cellcolor[HTML]{FFFFFF}PGD~\cite{madry2018towards}} &
  \multicolumn{2}{c}{\cellcolor[HTML]{FFFFFF}PAA~\cite{xu2023sok}} &
  \textbf{AdvGPS} \\ \midrule
   -&
   -&
  \cellcolor[HTML]{FFFFFF}W.b./B.b. &
  W.b./B.b. &
  W.b. &
  B.b. &
  W.b. &
  B.b. &
  W.b. &
  B.b. &
  W.b. &
  B.b. &
  B.b.\\ \midrule
No Fusion &
  \multicolumn{1}{c|}{\cellcolor[HTML]{FFFFFF}- } &
  \multicolumn{1}{c|}{\cellcolor[HTML]{FFFFFF}0.557} &
  \multicolumn{1}{c|}{\cellcolor[HTML]{FFFFFF}-} &
  - &
  \multicolumn{1}{c|}{\cellcolor[HTML]{FFFFFF}-} &
  - &
  \multicolumn{1}{c|}{\cellcolor[HTML]{FFFFFF}-} &
  - &
  \multicolumn{1}{c|}{\cellcolor[HTML]{FFFFFF}-} &
  - &
  \multicolumn{1}{c|}{\cellcolor[HTML]{FFFFFF}-} &
  - \\ \midrule
\cellcolor[HTML]{FFFFFF} &
  \multicolumn{1}{c|}{\cellcolor[HTML]{FFFFFF}$G_{xyz}$ } &
  \multicolumn{1}{c|}{\cellcolor[HTML]{FFFFFF}} &
  \multicolumn{1}{c|}{\cellcolor[HTML]{FFFFFF}0.657} &
  0.572 &
  \multicolumn{1}{c|}{\cellcolor[HTML]{FFFFFF}0.583} &
  0.626 &
  \multicolumn{1}{c|}{\cellcolor[HTML]{FFFFFF}0.638} &
  0.567 &
  \multicolumn{1}{c|}{\cellcolor[HTML]{FFFFFF}0.612} &
  {\color{blue}0.550} &
  \multicolumn{1}{c|}{\cellcolor[HTML]{FFFFFF}0.561} &
   \textbf{\color{red}0.502} \\
\multirow{-2}{*}{\cellcolor[HTML]{FFFFFF}Attfuse~\cite{xu2022opv2v}} &
  \multicolumn{1}{c|}{\cellcolor[HTML]{FFFFFF}$G_{all}$} &
  \multicolumn{1}{c|}{\multirow{-2}{*}{\cellcolor[HTML]{FFFFFF}0.918}} &
  \multicolumn{1}{c|}{\cellcolor[HTML]{FFFFFF}0.669} &
  0.571 &
  \multicolumn{1}{c|}{\cellcolor[HTML]{FFFFFF}0.579} &
  0.624 &
  \multicolumn{1}{c|}{\cellcolor[HTML]{FFFFFF}0.626} &
  0.572 &
  \multicolumn{1}{c|}{\cellcolor[HTML]{FFFFFF}0.597} &
  {\color{blue}0.540} &
  \multicolumn{1}{c|}{\cellcolor[HTML]{FFFFFF}0.559} &
   \textbf{\color{red}0.499} \\ \midrule
\cellcolor[HTML]{FFFFFF} &
  \multicolumn{1}{c|}{\cellcolor[HTML]{FFFFFF}$G_{xyz}$ } &
  \multicolumn{1}{c|}{\cellcolor[HTML]{FFFFFF}} &
  \multicolumn{1}{c|}{\cellcolor[HTML]{FFFFFF}0.526} &
  0.508 &
  \multicolumn{1}{c|}{\cellcolor[HTML]{FFFFFF}0.550} &
  0.443 &
  \multicolumn{1}{c|}{\cellcolor[HTML]{FFFFFF}0.601} &
  0.422 &
  \multicolumn{1}{c|}{\cellcolor[HTML]{FFFFFF}0.583} &
  {\color{blue}0.421} &
  \multicolumn{1}{c|}{\cellcolor[HTML]{FFFFFF}0.527} &
   \textbf{\color{red}0.299} \\
\multirow{-2}{*}{\cellcolor[HTML]{FFFFFF}F-Cooper~\cite{chen2019f}} &
  \multicolumn{1}{c|}{\cellcolor[HTML]{FFFFFF}$G_{all}$} &
  \multicolumn{1}{c|}{\multirow{-2}{*}{\cellcolor[HTML]{FFFFFF}0.898}} &
  \multicolumn{1}{c|}{\cellcolor[HTML]{FFFFFF}0.538} &
  0.510 &
  \multicolumn{1}{c|}{\cellcolor[HTML]{FFFFFF}0.549} &
  0.436 &
  \multicolumn{1}{c|}{\cellcolor[HTML]{FFFFFF}0.586} &
  0.419 &
  \multicolumn{1}{c|}{\cellcolor[HTML]{FFFFFF}0.562} &
  {\color{blue}0.413} &
  \multicolumn{1}{c|}{\cellcolor[HTML]{FFFFFF}0.524} &
   \textbf{\color{red}0.298} \\ \midrule
\cellcolor[HTML]{FFFFFF} &
  \multicolumn{1}{c|}{\cellcolor[HTML]{FFFFFF}$G_{xyz}$ } &
  \multicolumn{1}{c|}{\cellcolor[HTML]{FFFFFF}} &
  \multicolumn{1}{c|}{\cellcolor[HTML]{FFFFFF}0.568} &
  0.565 &
  \multicolumn{1}{c|}{\cellcolor[HTML]{FFFFFF}0.593} &
  0.523 &
  \multicolumn{1}{c|}{\cellcolor[HTML]{FFFFFF}0.653} &
  {\color{blue}0.516} &
  \multicolumn{1}{c|}{\cellcolor[HTML]{FFFFFF}0.628} &
  0.523 &
  \multicolumn{1}{c|}{\cellcolor[HTML]{FFFFFF}0.563} &
   \textbf{\color{red}0.422} \\
\multirow{-2}{*}{\cellcolor[HTML]{FFFFFF}V2VAM~\cite{10077757}} &
  \multicolumn{1}{c|}{\cellcolor[HTML]{FFFFFF}$G_{all}$} &
  \multicolumn{1}{c|}{\multirow{-2}{*}{\cellcolor[HTML]{FFFFFF}0.920}} &
  \multicolumn{1}{c|}{\cellcolor[HTML]{FFFFFF}0.584} &
  0.566 &
  \multicolumn{1}{c|}{\cellcolor[HTML]{FFFFFF}0.592} &
  0.523 &
  \multicolumn{1}{c|}{\cellcolor[HTML]{FFFFFF}0.638} &
  {\color{blue}0.513} &
  \multicolumn{1}{c|}{\cellcolor[HTML]{FFFFFF}0.607} &
  0.520 &
  \multicolumn{1}{c|}{\cellcolor[HTML]{FFFFFF}0.557} &
   \textbf{\color{red}0.433} \\ \midrule
\cellcolor[HTML]{FFFFFF} &
  \multicolumn{1}{c|}{\cellcolor[HTML]{FFFFFF}$G_{xyz}$ } &
  \multicolumn{1}{c|}{\cellcolor[HTML]{FFFFFF}} &
  \multicolumn{1}{c|}{\cellcolor[HTML]{FFFFFF}0.589} &
  0.599 &
  \multicolumn{1}{c|}{\cellcolor[HTML]{FFFFFF}0.615} &
  0.573 &
  \multicolumn{1}{c|}{\cellcolor[HTML]{FFFFFF}0.671} &
  0.546 &
  \multicolumn{1}{c|}{\cellcolor[HTML]{FFFFFF}0.644} &
  {\color{blue}0.533} &
  \multicolumn{1}{c|}{\cellcolor[HTML]{FFFFFF}{\color{blue}0.533}} &
   \textbf{\color{red}0.433} \\
\multirow{-2}{*}{\cellcolor[HTML]{FFFFFF}V2X-ViT~\cite{xu2022v2x}} &
  \multicolumn{1}{c|}{\cellcolor[HTML]{FFFFFF}$G_{all}$} &
  \multicolumn{1}{c|}{\multirow{-2}{*}{\cellcolor[HTML]{FFFFFF}0.928}} &
  \multicolumn{1}{c|}{\cellcolor[HTML]{FFFFFF}0.597} &
  0.600 &
  \multicolumn{1}{c|}{\cellcolor[HTML]{FFFFFF}0.615} &
  0.575 &
  \multicolumn{1}{c|}{\cellcolor[HTML]{FFFFFF}0.658} &
  0.549 &
  \multicolumn{1}{c|}{\cellcolor[HTML]{FFFFFF}0.628} &
  {\color{blue}0.529} &
  \multicolumn{1}{c|}{\cellcolor[HTML]{FFFFFF}{\color{blue}0.529}} &
   \textbf{\color{red}0.434} \\ \midrule
\cellcolor[HTML]{FFFFFF} &
  \multicolumn{1}{c|}{\cellcolor[HTML]{FFFFFF}$G_{xyz}$ } &
  \multicolumn{1}{c|}{\cellcolor[HTML]{FFFFFF}} &
  \multicolumn{1}{c|}{\cellcolor[HTML]{FFFFFF}0.570} &
  0.570 &
  \multicolumn{1}{c|}{\cellcolor[HTML]{FFFFFF}0.592} &
  0.547 &
  \multicolumn{1}{c|}{\cellcolor[HTML]{FFFFFF}0.649} &
  0.517 &
  \multicolumn{1}{c|}{\cellcolor[HTML]{FFFFFF}0.624} &
  {\color{blue}0.498} &
  \multicolumn{1}{c|}{\cellcolor[HTML]{FFFFFF}0.559} &
   \textbf{\color{red}0.397} \\
\multirow{-2}{*}{\cellcolor[HTML]{FFFFFF}CoBVET~\cite{xu2022cobevt}} &
  \multicolumn{1}{c|}{\cellcolor[HTML]{FFFFFF}$G_{all}$} &
  \multicolumn{1}{c|}{\multirow{-2}{*}{\cellcolor[HTML]{FFFFFF}0.904}} &
   \multicolumn{1}{c|}{\cellcolor[HTML]{FFFFFF}0.577} &
   0.568 &
   \multicolumn{1}{c|}{\cellcolor[HTML]{FFFFFF}0.592} &
   0.545 &
   \multicolumn{1}{c|}{\cellcolor[HTML]{FFFFFF}0.634} &
   0.520 &
   \multicolumn{1}{c|}{\cellcolor[HTML]{FFFFFF}0.610} &
    {\color{blue}0.493}&
   \multicolumn{1}{c|}{\cellcolor[HTML]{FFFFFF}0.553} &
   \textbf{\color{red}0.403} \\ \bottomrule
\end{tabular}%
}
\end{table*}

\begin{figure*}[!t]
\centering
% \captionsetup[subfloat]{labelsep=none,format=plain,labelformat=empty}
\subfloat[No Attack~\cite{chen2019f}]{%
  \includegraphics[width=0.63\columnwidth]{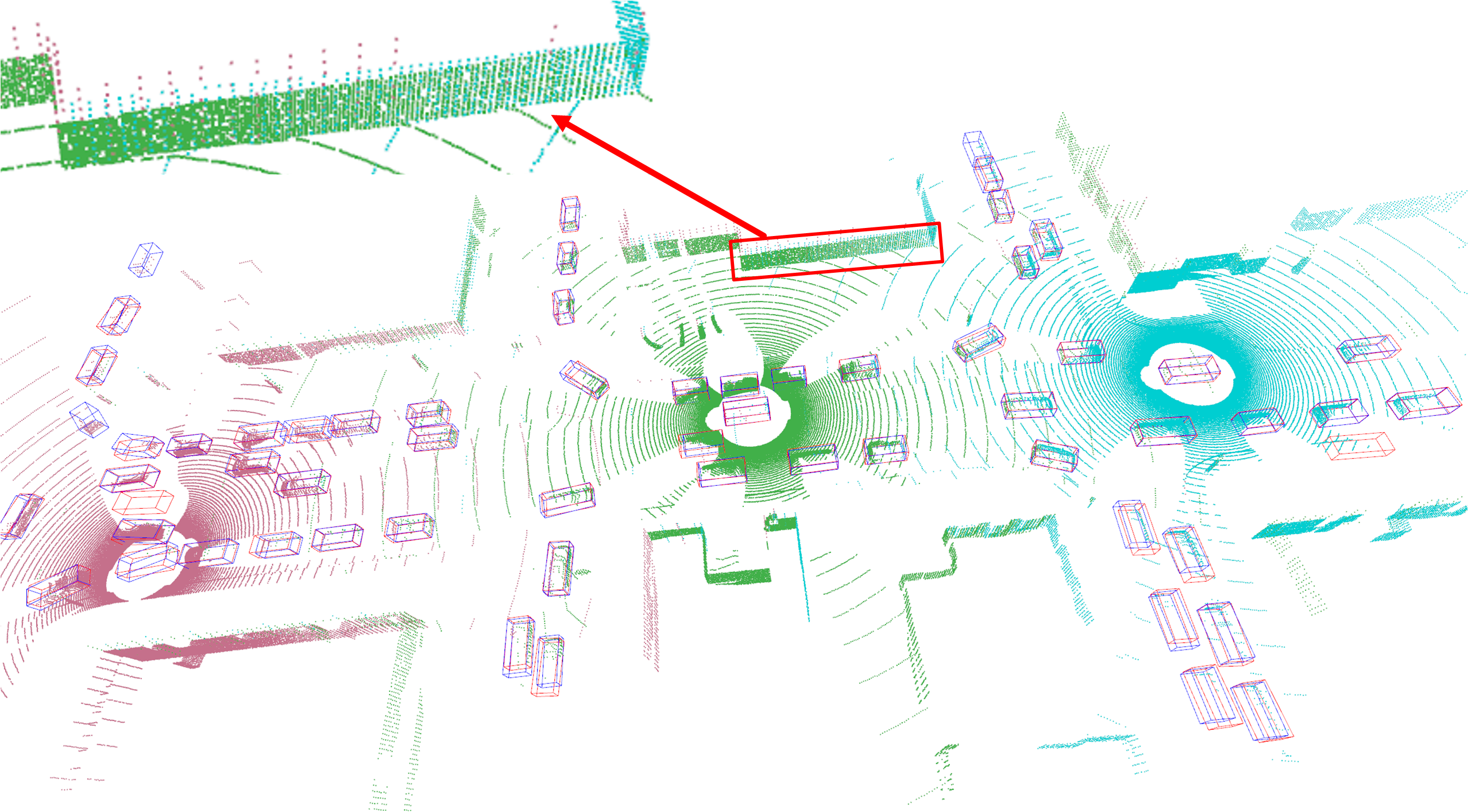}%
}
\hfil
\subfloat[RBA~\cite{yang2023anomaly}]{%
  \includegraphics[width=0.63\columnwidth]{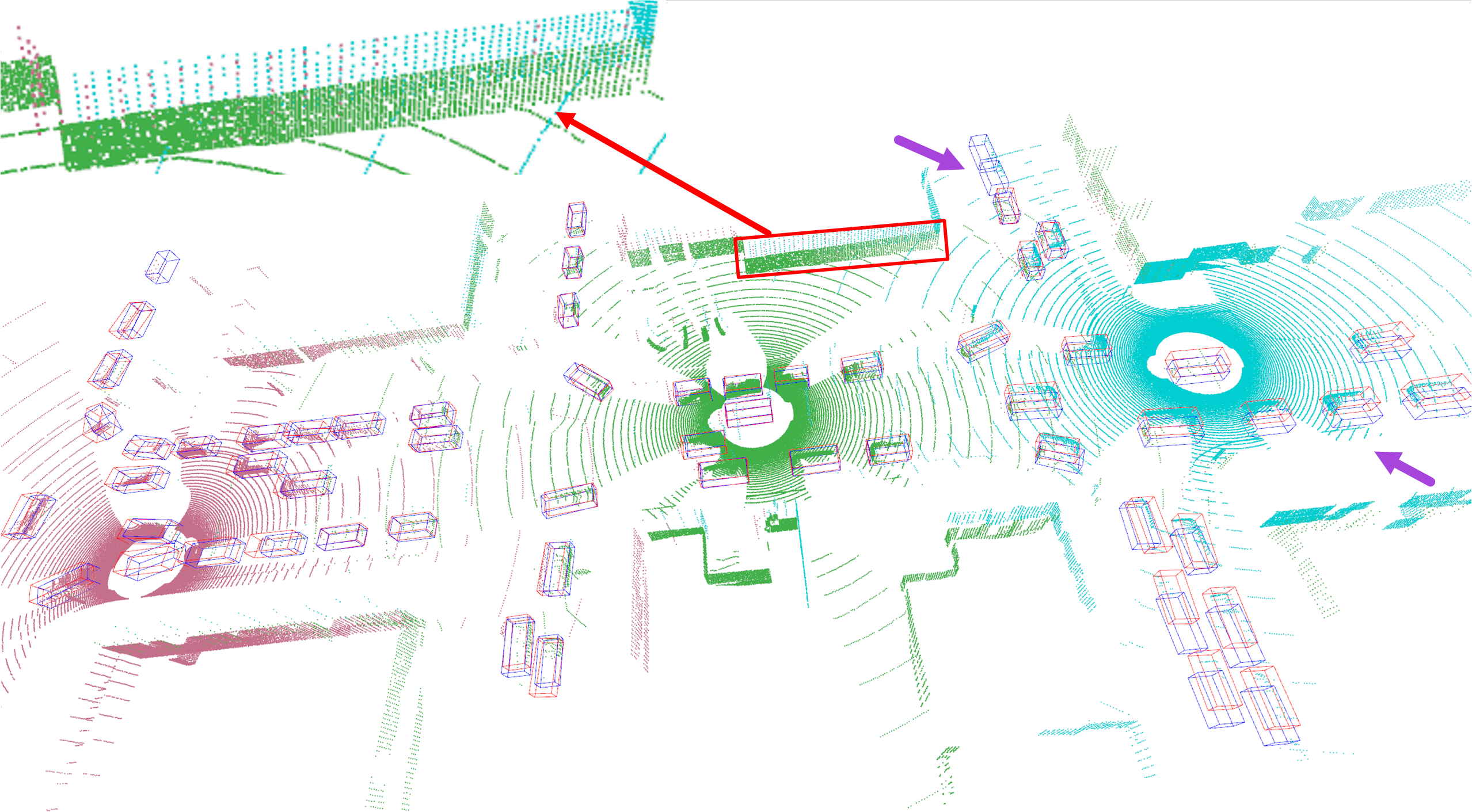}%
}
\hfil
\subfloat[IFSGM~\cite{kurakin2018adversarial}]{%
  \includegraphics[width=0.63\columnwidth]{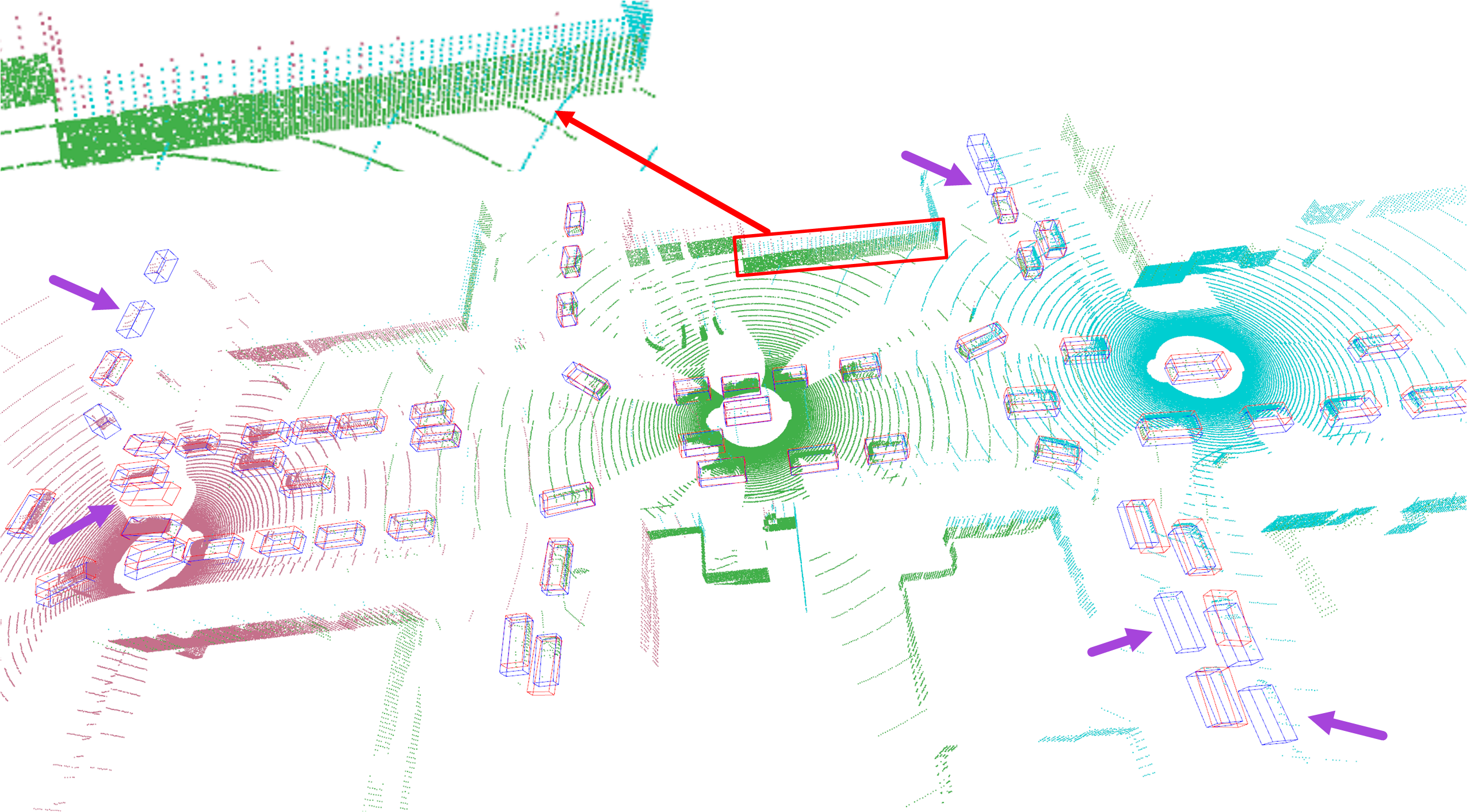}%
}
% \hfil
\\
\subfloat[PGD~\cite{madry2018towards}]{%
  \includegraphics[width=0.63\columnwidth]{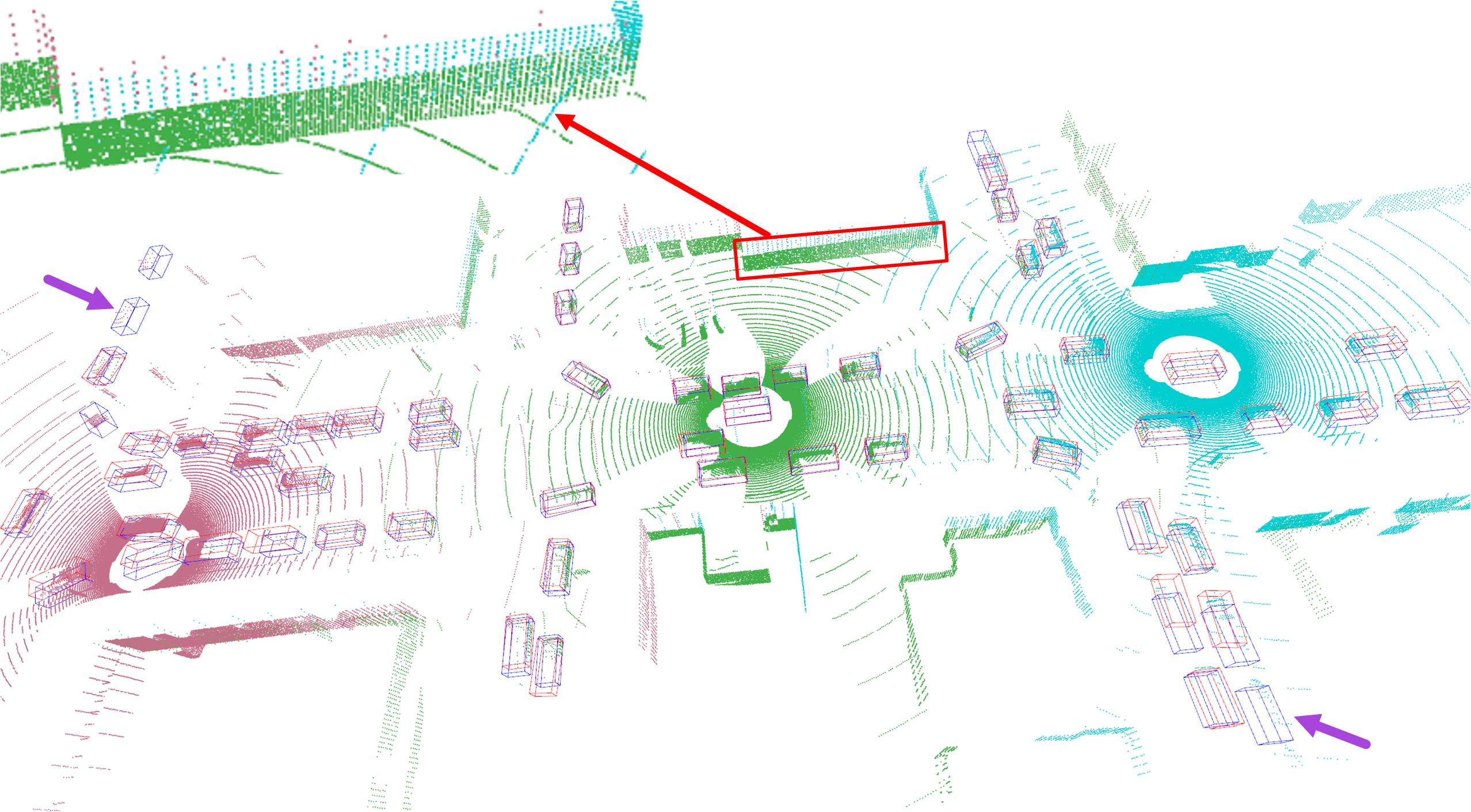}%
}
\hfil
\subfloat[PAA~\cite{xu2022v2x}]{%
  \includegraphics[width=0.63\columnwidth]{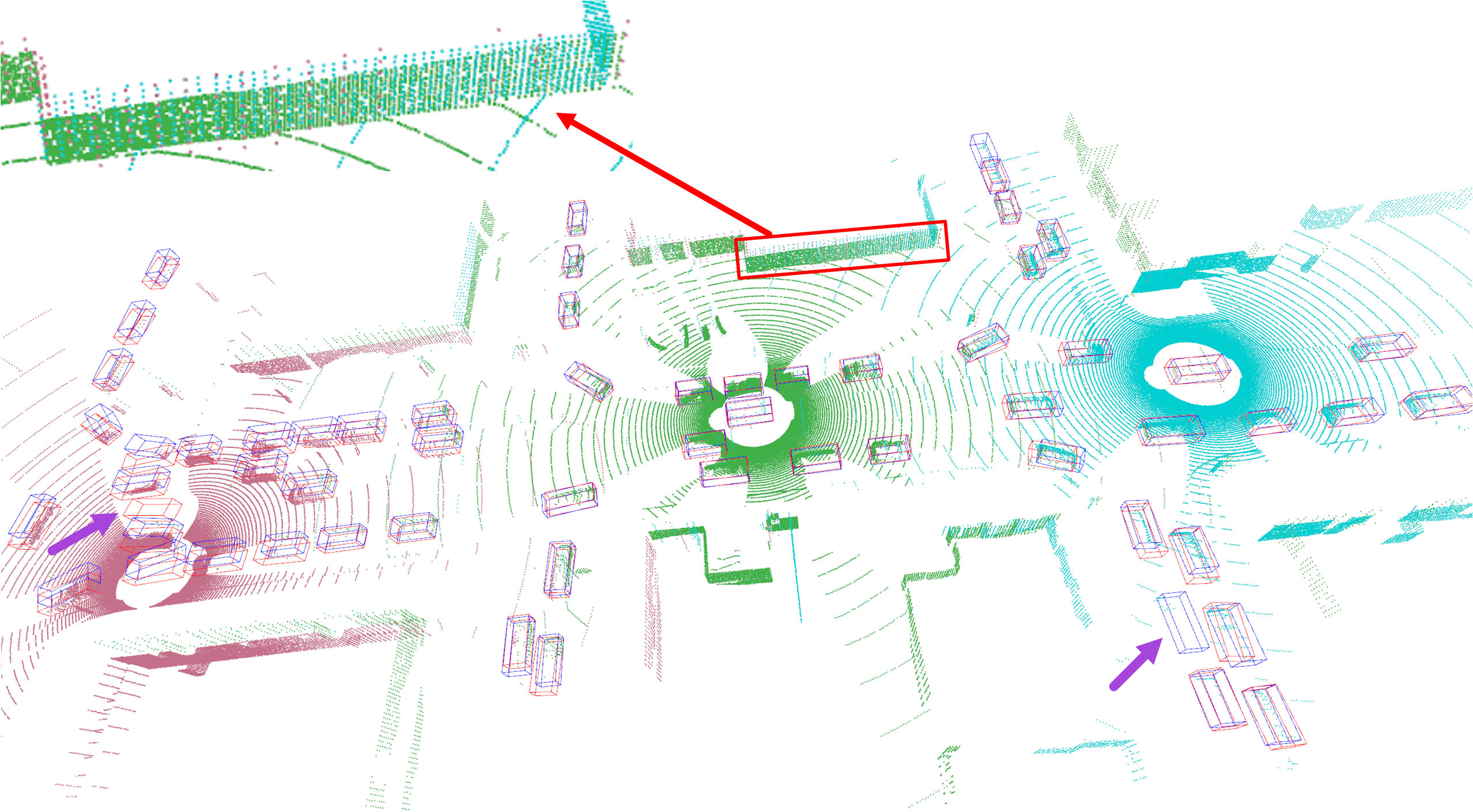}%
}
\hfil
\subfloat[AdvGPS]{%
  \includegraphics[width=0.63\columnwidth]{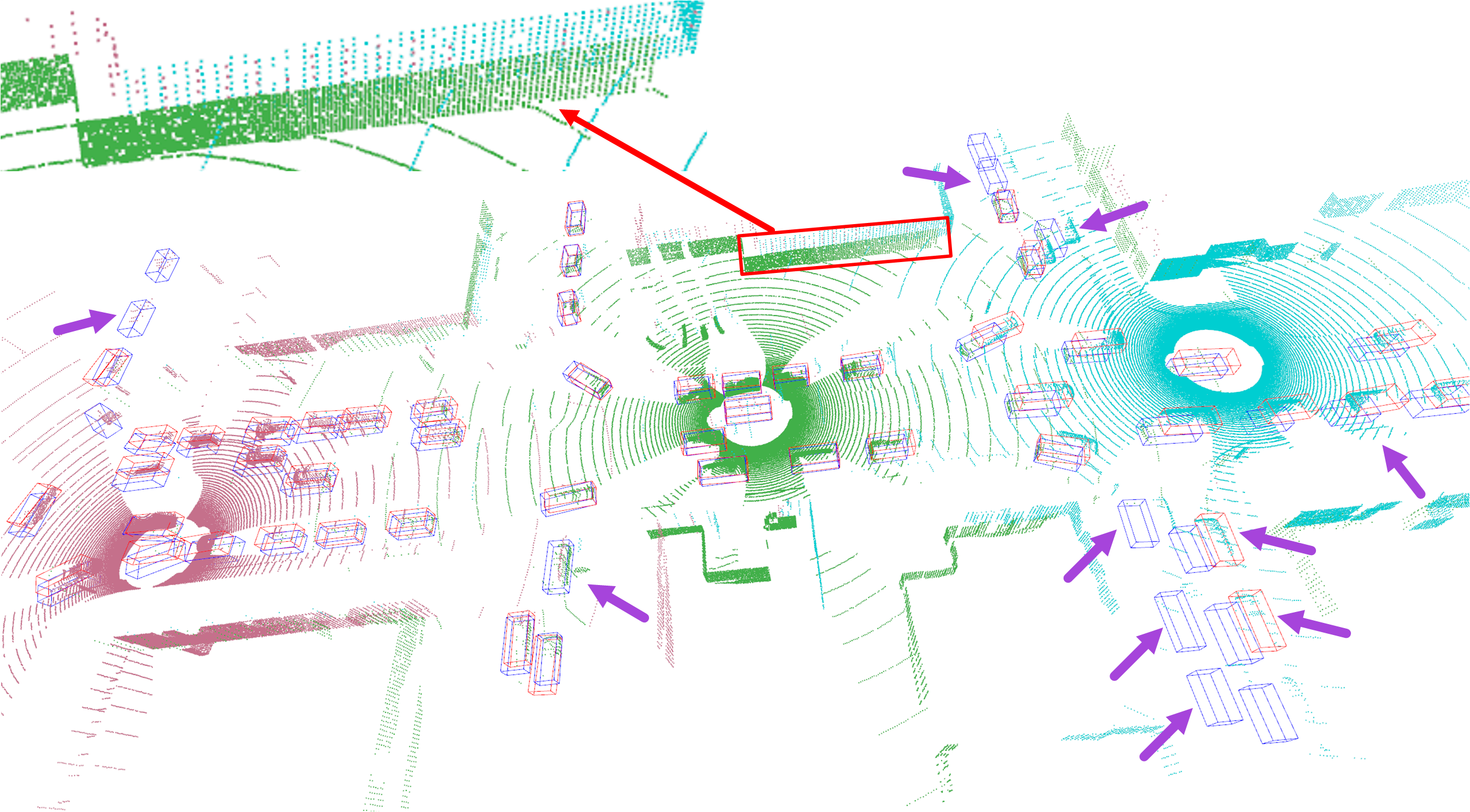}%
}
\caption{\textbf{3D detection visualization on attacking V2V  model CoBEVT~\cite{xu2022cobevt}}. \textcolor{blue}{Blue} and \textcolor{amaranth}{red} 3D bounding boxes represent \textcolor{blue}{ground truth} and \textcolor{amaranth}{prediction}. Purple arrows: detection errors. {\color{ao(english)} Green} point cloud: by ego vehicle. Other-color point clouds: projected to ego coordinate by nearby attacked CAVs.} 
\label{fig:visualization}
\vspace{-10pt}
\end{figure*}

\subsection{Experimental Setups}

% \subsubsection{Dataset}
\noindent \textbf{Dataset:}
\label{subsubsec:dataset}
We conduct GPS attack experiments on the publicized OPV2V dataset~\cite{xu2022opv2v} for V2V cooperative perception tasks. OPV2V~\cite{xu2022opv2v} is a publicized open-source simulated dataset for V2V perception, which is collected in CARLA and OpenCDA \cite{xu2021opencda}. Following the default setting of OPV2V \cite{xu2022opv2v}, we use $594$ frames in the digital town of Culver City, Los Angeles with the same road topology used as the testing set for all methods.
% we use $6764$ frames and $1,981$ frames from OPV2V as the training set and validation set, respectively

\noindent \textbf{3D detection methods:}
We selected the five state-of-the-art cooperative perception methods as the point cloud based 3D detection models, \ie,  AttFuse~\cite{xu2022opv2v}, F-cooper~\cite{chen2019f}, V2VAM~\cite{10077757}, V2X-ViT~\cite{xu2022v2x} and CoBEVT~\cite{xu2022cobevt}. All methods leverage the anchor-based PointPillar method~\cite{lang2019pointpillars} to extract visual features from point clouds because of its low inference latency and optimized memory usage.

\noindent \textbf{Evaluation metrics:} 
% We evaluate the performance of our proposed framework by the final 3D vehicle detection accuracy. Following~\cite{xu2022opv2v,xu2022v2x}, we set the evaluation range as $x\in[-140, 140]$ meters,  $y\in[-40, 40]$ meters, where all CAVs are included in this spatial range, whose number is in the range of $[1,5]$ in the experiment. We measure the accuracy with Average Precision (AP) at the Intersection-over-Union (IoU) threshold of 0.5.
We assess the final 3D vehicle detection accuracy for our proposed framework performance. Similar to prior works~\cite{xu2022opv2v,xu2022v2x}, we set $x\in[-140, 140]$ meters,  $y\in[-40, 40]$ meters as the evaluation range, where all CAVs (in the range of $[1,5]$) are included in this spatial range. We employ Average Precision (AP) at the Intersection-over-Union (IoU) threshold of 0.5 as the accuracy metric.

\noindent\textbf{Attack settings:}
To evaluate the effectiveness of our AdvGPS attack on the V2V perception pipeline, we compare it with three commonly used attack methods on CAVs' GPS poses: Projected Gradient Descent (PGD) \cite{madry2018towards}, Fast Gradient Sign Method (FGSM) \cite{43405}, and Iterative Fast Gradient Sign Method (IFGSM) \cite{kurakin2018adversarial}. Additionally, we include two current GPS spoofing methods: Random Bias Attack (RBA) \cite{yang2023anomaly} and Position Altering Attack (PAA) \cite{xu2023sok}. 
FGSM, IFGSM, PGD, and PAA are gradient-based attack methods, following the \textit{same} implementation settings described in Section~\ref{subsec:impl}. RBA can be implemented by applying uniformly distributed random bias within the same range settings ($\epsilon_{x, y}$, $\epsilon_{z}$ and $\epsilon_{\theta_{x}, \theta_{y}, \theta_{z}}$) as in Section~\ref{subsec:impl}. We use the Adam optimizer~\cite{loshchilov2017decoupled}, and all models are executed on two RTX 3090 GPUs.
We conduct both black-box and white-box attacks to assess the vulnerability and transferability of the cooperative perception methods. These attack scenarios are described as: 
\begin{itemize}
    \item \textit{White-Box Attack (W.b.)}: All attack methods utilize PointPillar~\cite{lang2019pointpillars} as the encoder and AttFuse~\cite{xu2022opv2v} as the feature fusion network. The adversarial GPS information of all attacked CAVs for the entire duration of the frames is saved to be used for attacking other cooperative perception methods.
    
    \item \textit{Black-Box Attack (B.b.)}: We employ VoxelNet~\cite{zhou2018voxelnet} as the encoder and a naive self-attention model as the feature fusion network in training. We save the   adversarial GPS information of all attacked CAVs across the entire frame duration to be used for attacking the \textit{unseen}  PointPillar~\cite{lang2019pointpillars} encoder and \textit{other} state-of-the-art fusion methods of cooperative perception.
\end{itemize}

\subsection{Quantitative Evaluation}

\noindent\textbf{GPS attack performance analysis:}
Table~\ref{tab:gps_result} overviews the GPS attack results on the 3D object detection task. We conducted attacks on three parameters $[x, y, z]$ and all six parameters $[x, y, z, \theta_{x}, \theta_{y}, \theta_{z}]$ of the CAV poses $G_{cav}$, denoted as $G_{xyz}$ and $G_{all}$, respectively.
In the \textit{White-Box Attack} scenario, all attack methods demonstrated the ability to degrade the detection performance of state-of-the-art (SOTA) intermediate fusion methods. For instance, the traditional random bias method RBA~\cite{yang2023anomaly} achieved $58.9\%$ for $G_{xyz}$ and $59.7\%$ for $G_{all}$ when attacking the V2X-ViT~\cite{xu2022v2x} model. In contrast, the gradient-based PAA~\cite{xu2023sok} achieved $53.3\%$ for $G_{xyz}$ and $52.9\%$ for $G_{all}$, both of which reduced the performance below that of \textit{No Fusion}. This indicates that GPS attacks can easily undermine the advantages of collaboration among intermediate fusion methods.
In the \textit{Black-Box Attack} scenario, AdvGPS outperforms all other attacks across various SOTA fusion methods (See Table~\ref{tab:gps_result}). When attacking the $G_{xyz}$ of CAV poses, AdvGPS achieved $29.9\%$ on F-Cooper~\cite{chen2019f}, $42.2\%$ on V2VAM~\cite{10077757}, $43.3\%$ on V2X-ViT~\cite{xu2022v2x}, and $39.7\%$ on CoBEVT~\cite{xu2022cobevt}. These results exceeded the second-best attacks by approximately $12.1\%$, $9.4\%$, $10.0\%$, and $10.1\%$, respectively. Our AdvGPS, leveraging three types of statistically sensitive natural discrepancies, \ie, appearance-based discrepancy, distribution-based discrepancy, and task-aware discrepancy, effectively reduces object detection accuracy and demonstrates remarkable vulnerability and transferability in multi-agent perception tasks.
Furthermore, when comparing the results of the \textit{Black-Box Attack} setting for AdvGPS with the \textit{White-Box Attack} results for all five attacks, AdvGPS consistently achieves the highest attack success rate. Comparing the results of attacking all six parameters $G_{all}$ and only $G_{xyz}$ reveals their similar performance, which indicates  that directly attacking the position of CAVs' poses $G_{xyz}$ is an efficient approach to reduce computation complexity while maintaining exceptional attack performance in real-world scenarios.

% \noindent\textbf{Vulnerable GPS parameters Analysis:} We investigate the  sensitivity of each parameter of CAV poses $p_{cav}^{t}$, as shown in Table~\ref{tab:variousGPS}. 
% Within the realm of actual GPS errors, attacking each parameter based on PAA and our proposed AdvGPS is still efficient in reducing object detection accuracy for two ViT-based models, \textit{i.e.}, V2X-ViT~\cite{} and CoBEVT~\cite{}.
% Intuitively, within the constraints of a small range of GPS pose changes, attacking the position parameters $x$ and $y$ of CAVs' poses is the more efficient approach than other four parameters, indicating the sensitive of V2V cooperative perception system.
% In addition,  we set the position ($(x, y, z)$) as the max value of GPS pose constraint to attack Attfuse~\cite{}, and obtains $65.1\%$ in AP=0.5. It is very close to attack by SBA~\cite{} method.

\noindent\textbf{Sensitivity analysis of GPS parameters:}
We have conducted a sensitivity analysis for each parameter of CAV poses $\widehat{G}_{cav}$ in Table~\ref{tab:variousGPS}. 
Even when operating within the boundaries of typical GPS errors, both the PAA~\cite{xu2023sok} and our AdvGPS methods remain effective in significantly reducing object detection accuracy for two ViT-based models, namely V2X-ViT~\cite{xu2022v2x} and CoBEVT~\cite{xu2022cobevt}.
Intuitively, among all six parameters, attacking the positional parameters $x$ and $y$ of CAVs' poses is the most efficient approach within a small range of GPS pose changes, highlighting their sensitivity within the V2V cooperative perception system.
Furthermore, when we set the positional parameters $[x, y, z]$ to have maximum GPS bias $[1.118, 1.118, 1.395]$ meters of the actual measured GPS errors in real world~\cite{chiang2020performance} and attack Attfuse~\cite{xu2022opv2v}, we obtained a AP $65.1\%$ at IoU 0.5, close to the performance by RBA~\cite{yang2023anomaly}, which has less degradation than our AdvGPS. It demonstrates that the gradient optimization based attack methods are  better than the non-gradient attack.   

% \noindent\textbf{Ablation Study:} 
% As Table~\ref{tab:abla} depicts, all three components in our AdvGPS attack framework have contributed to more degradation of detection performance on CoBVET~\cite{xu2022cobevt}. Under the setting of \textit{Black Box Attack}, adding the $D_{appe}$, $D_{dist}$, $D_{task}$ can degrade the performance by $48.1\%$, $037.2\%$, and $027.0\%$, respectively, indicating our design's efficiency.
% %
% Obviously, our proposed AdvGPS attacks substantially under-mine the performance of state-of-the-art V2V cooperative perception methods.

\noindent\textbf{Ablation study:} 
As shown in Table~\ref{tab:abla}, our comprehensive ablation study reveals that all three components within our AdvGPS attack framework contribute significantly to the degradation of detection performance on CoBVET~\cite{xu2022cobevt}. In the context of the \textit{Black-Box Attack} setting, the inclusion of $D_{app}$, $D_{dist}$, and $D_{task}$ results in performance degradation by $48.1\%$, $37.2\%$, and $27.0\%$, respectively. This nuanced analysis emphasizes the effectiveness of our design. It is evident that our proposed AdvGPS attacks significantly undermine the performance of state-of-the-art V2V cooperative perception methods.

% \noindent\textbf{3D detection visualization:}
% We visually compare different attack methods under \textit{Black Box Attack} settings and show the attack result on CoBEVT~\cite{} in Fig.~\ref{fig:visualization}.
% %
% Intuitively, these all comparison attack methods can degrade the detection performance successfully, thus leading to some false negative proposals.
% %
% While the proposed method achieve the best attack success rate  with more false negative proposals, which is highlighted in~\ref{fig:visualization} .
% %
% In addition, the other CAVs' attacked projected colorful point cloud under ego's coordinate, and ego's green point cloud are zoomed to show details of ~\ref{fig:visualization}. 

\noindent\textbf{3D detection visualization:}
Under the \textit{Black-Box Attack} setting, we compare the visualizations of different attacks and showcase their impact on CoBEVT~\cite{xu2022cobevt} in Fig.~\ref{fig:visualization}. It is noteworthy that our proposed method achieves the best attack success rate with a notable increase in false-negative and false-positive detection errors, as highlighted in Fig.~\ref{fig:visualization}.
Additionally, as shown in Fig.~\ref{fig:visualization}, we project the attacked point clouds (other colors) from nearby CAVs to the coordinate system of ego vehicle (green color), then we can  discover that the attacked point clouds by our AdvGPS has  small point cloud shift (similar to random bias), which verifies that our AdvGPS attack is stealthy.

% \usepackage{booktabs}
% \usepackage{multirow}
% \usepackage{graphicx}
\begin{table}[]
\caption{\textbf{Attack results of various GPS parameters  under the setting of \textit{Black-Box Attack}.} We show the Average Precision (AP) at IoU=0.5.}
\label{tab:variousGPS}
\resizebox{\columnwidth}{!}{%
\begin{tabular}{@{}
>{\columncolor[HTML]{FFFFFF}}c 
>{\columncolor[HTML]{FFFFFF}}c 
>{\columncolor[HTML]{FFFFFF}}c 
>{\columncolor[HTML]{FFFFFF}}c 
>{\columncolor[HTML]{FFFFFF}}c 
>{\columncolor[HTML]{FFFFFF}}c 
>{\columncolor[HTML]{FFFFFF}}c 
>{\columncolor[HTML]{FFFFFF}}c @{}}
\toprule
Model                                           & Attack                                                & $\hat{x}$  & $\hat{y}$  & $\hat{z}$  & $\hat{\theta}_{x}$ & $\hat{\theta}_{y}$ & $\hat{\theta}_{z}$ \\ \midrule
\cellcolor[HTML]{FFFFFF}                         & \multicolumn{1}{c|}{\cellcolor[HTML]{FFFFFF}PAA~\cite{xu2023sok}} & 0.676 & 0.615 & 0.783 &\textbf{\color{red}0.802}    & 0.802  & \textbf{\color{red}0.801}      \\
\multirow{-2}{*}{\cellcolor[HTML]{FFFFFF}V2X-ViT~\cite{xu2022v2x}}    & \multicolumn{1}{c|}{\cellcolor[HTML]{FFFFFF}\textbf{AdvGPS}}  & \textbf{\color{red}0.453 } & \textbf{\color{red}0.396 }  & \textbf{\color{red}0.737} & \textbf{\color{red}0.802}    & \textbf{\color{red}0.800}   & \textbf{\color{red}0.801}     \\ \midrule
\cellcolor[HTML]{FFFFFF}                         & \multicolumn{1}{c|}{\cellcolor[HTML]{FFFFFF}PAA~\cite{xu2023sok}} & 0.664 & 0.609 & 0.759 & \textbf{\color{red}0.784}   & 0.781  & \textbf{\color{red}0.783}    \\
\multirow{-2}{*}{\cellcolor[HTML]{FFFFFF}CoBEVT~\cite{xu2022cobevt}} & \multicolumn{1}{c|}{\cellcolor[HTML]{FFFFFF}\textbf{AdvGPS}}  & \textbf{\color{red}0.450}  & \textbf{\color{red}0.377} & \textbf{\color{red}0.732} & 0.785   & \textbf{\color{red}0.780}  & \textbf{\color{red}0.783}    \\ \bottomrule
\end{tabular}%
} \vspace{-10pt}
\end{table}

\begin{table}[]
\centering
\caption{\textbf{Ablation study of proposed AdvGPS attack under the setting of  \textit{Black-Box Attack}.} The six parameters of CAV GPS pose $G_{all}$ are used to  attack CoBVET~\cite{xu2022cobevt}.}
\tiny
\centering
\resizebox{0.8\columnwidth}{!}{%
\begin{tabular}{cccc}
\toprule
$D_{app}$ & $D_{dist}$ & $D_{task}$ & AP@IoU=0.5 \\ \midrule

           &             &           & 0.904 (No Attack)     \\  
\checkmark &             &           & 0.423 (\textbf{\color{gray} -0.481})       \\
           & \checkmark  &           &  0.532 (\textbf{\color{gray} -0.372})        \\
          &              & \checkmark &  0.634 (\textbf{\color{gray} -0.270})        \\
\checkmark & \checkmark  &  \checkmark & 0.403 (\textbf{\color{gray} -0.501})   \\ \bottomrule
\end{tabular}
    }
\label{tab:abla}
\vspace{-10pt}
\end{table}

\section{Conclusions}\label{Sec:Conclusions}
% This paper is the first work that investigates the GPS attack directly on multi-agent cooperation perception.
% this paper pioneers the investigation of GPS attacks on multi-agent cooperative perception systems. Our AdvGPS framework, designed to subtly perturb CAVs' GPS signals, reveals the vulnerabilities of cooperative perception in real-world scenarios. Through extensive experiments on the OPV2V dataset, we demonstrate the effectiveness and transferability of AdvGPS attacks. This work emphasizes the critical need to enhance the security of multi-agent cooperative perception systems, a crucial consideration for the autonomous vehicle industry.
This paper pioneers the investigation of adversarial GPS attacks within the multi-agent cooperative perception systems. Introducing \textbf{AdvGPS}, our method generates subtle yet effective adversarial GPS signals that stealthily mislead individual agents, leading to a significant reduction in object detection accuracy. To bolster the effectiveness of the AdvGPS attacks in black-box scenarios, we introduce three statistically sensitive natural discrepancies: appearance-based, distribution-based, and task-aware discrepancies.
Extensive experiments conducted on the OPV2V dataset highlight the substantial performance degradation caused by our AdvGPS attacks across various point cloud-based 3D detection systems. This groundbreaking discovery underscores the pressing need to address security concerns within multi-agent perception systems, marking a critical frontier in research.
\bibliographystyle{IEEEtran}
\bibliography{Jinlong}

\begin{thebibliography}{10}
\providecommand{\url}[1]{#1}
\csname url@samestyle\endcsname
\providecommand{\newblock}{\relax}
\providecommand{\bibinfo}[2]{#2}
\providecommand{\BIBentrySTDinterwordspacing}{\spaceskip=0pt\relax}
\providecommand{\BIBentryALTinterwordstretchfactor}{4}
\providecommand{\BIBentryALTinterwordspacing}{\spaceskip=\fontdimen2\font plus
\BIBentryALTinterwordstretchfactor\fontdimen3\font minus
  \fontdimen4\font\relax}
\providecommand{\BIBforeignlanguage}[2]{{%
\expandafter\ifx\csname l@#1\endcsname\relax
\typeout{** WARNING: IEEEtran.bst: No hyphenation pattern has been}%
\typeout{** loaded for the language `#1'. Using the pattern for}%
\typeout{** the default language instead.}%
\else
\language=\csname l@#1\endcsname
\fi
#2}}
\providecommand{\BIBdecl}{\relax}
\BIBdecl

\bibitem{xu2022opv2v}
R.~Xu, H.~Xiang, X.~Xia, X.~Han, J.~Li, and J.~Ma, ``Opv2v: An open benchmark
  dataset and fusion pipeline for perception with vehicle-to-vehicle
  communication,'' in \emph{International Conference on Robotics and
  Automation}.\hskip 1em plus 0.5em minus 0.4em\relax IEEE, 2022, pp.
  2583--2589.

\bibitem{xu2023v2v4real}
R.~Xu, X.~Xia, J.~Li, H.~Li, S.~Zhang, Z.~Tu, Z.~Meng, H.~Xiang, X.~Dong,
  R.~Song \emph{et~al.}, ``V2v4real: A real-world large-scale dataset for
  vehicle-to-vehicle cooperative perception,'' in \emph{IEEE/CVF Conference on
  Computer Vision and Pattern Recognition}, 2023, pp. 13\,712--13\,722.

\bibitem{xu2021opencda}
R.~Xu, Y.~Guo, X.~Han, X.~Xia, H.~Xiang, and J.~Ma, ``Opencda: an open
  cooperative driving automation framework integrated with co-simulation,'' in
  \emph{IEEE International Intelligent Transportation Systems
  Conference}.\hskip 1em plus 0.5em minus 0.4em\relax IEEE, 2021, pp.
  1155--1162.

\bibitem{li2023learning}
J.~Li, R.~Xu, X.~Liu, J.~Ma, Z.~Chi, J.~Ma, and H.~Yu, ``Learning for
  vehicle-to-vehicle cooperative perception under lossy communication,''
  \emph{IEEE Transactions on Intelligent Vehicles}, 2023.

\bibitem{yang2023anomaly}
Z.~Yang, J.~Ying, J.~Shen, Y.~Feng, Q.~A. Chen, Z.~M. Mao, and H.~X. Liu,
  ``Anomaly detection against gps spoofing attacks on connected and autonomous
  vehicles using learning from demonstration,'' \emph{IEEE Transactions on
  Intelligent Transportation Systems}, 2023.

\bibitem{xu2022v2x}
R.~Xu, H.~Xiang, Z.~Tu, X.~Xia, M.-H. Yang, and J.~Ma, ``V2x-vit:
  Vehicle-to-everything cooperative perception with vision transformer,'' in
  \emph{European Conference on Computer Vision}.\hskip 1em plus 0.5em minus
  0.4em\relax Springer, 2022, pp. 107--124.

\bibitem{10077757}
J.~Li, R.~Xu, X.~Liu, J.~Ma, Z.~Chi, J.~Ma, and H.~Yu, ``Learning for
  vehicle-to-vehicle cooperative perception under lossy communication,''
  \emph{IEEE Transactions on Intelligent Vehicles}, vol.~8, no.~4, pp.
  2650--2660, 2023.

\bibitem{xu2022cobevt}
R.~Xu, Z.~Tu, H.~Xiang, W.~Shao, B.~Zhou, and J.~Ma, ``Cobevt: Cooperative
  bird's eye view semantic segmentation with sparse transformers,'' in
  \emph{Conference on Robot Learning}, 2022.

\bibitem{xu2022bridging}
R.~Xu, J.~Li, X.~Dong, H.~Yu, and J.~Ma, ``Bridging the domain gap for
  multi-agent perception,'' in \emph{International Conference on Robotics and
  Automation}, 2023.

\bibitem{vadivelu2021learning}
N.~Vadivelu, M.~Ren, J.~Tu, J.~Wang, and R.~Urtasun, ``Learning to communicate
  and correct pose errors,'' in \emph{Conference on Robot Learning}.\hskip 1em
  plus 0.5em minus 0.4em\relax PMLR, 2021, pp. 1195--1210.

\bibitem{xu2023bridging}
R.~Xu, J.~Li, X.~Dong, H.~Yu, and J.~Ma, ``Bridging the domain gap for
  multi-agent perception,'' in \emph{IEEE International Conference on Robotics
  and Automation}.\hskip 1em plus 0.5em minus 0.4em\relax IEEE, 2023, pp.
  6035--6042.

\bibitem{hou2023evading}
Y.~Hou, Q.~Guo, Y.~Huang, X.~Xie, L.~Ma, and J.~Zhao, ``Evading deepfake
  detectors via adversarial statistical consistency,'' in \emph{IEEE/CVF
  Conference on Computer Vision and Pattern Recognition}, 2023, pp.
  12\,271--12\,280.

\bibitem{carlini2017towards}
N.~Carlini and D.~Wagner, ``Towards evaluating the robustness of neural
  networks,'' in \emph{IEEE Symposium on Security and Privacy}.\hskip 1em plus
  0.5em minus 0.4em\relax IEEE, 2017, pp. 39--57.

\bibitem{li2021fooling}
Y.~Li, C.~Wen, F.~Juefei-Xu, and C.~Feng, ``Fooling lidar perception via
  adversarial trajectory perturbation,'' in \emph{IEEE/CVF International
  Conference on Computer Vision}, 2021, pp. 7898--7907.

\bibitem{cheng2019improving}
S.~Cheng, Y.~Dong, T.~Pang, H.~Su, and J.~Zhu, ``Improving black-box
  adversarial attacks with a transfer-based prior,'' \emph{Advances in Neural
  Information Processing Systems}, vol.~32, 2019.

\bibitem{shi2019curls}
Y.~Shi, S.~Wang, and Y.~Han, ``Curls \& whey: Boosting black-box adversarial
  attacks,'' in \emph{IEEE/CVF Conference on Computer Vision and Pattern
  Recognition}, 2019, pp. 6519--6527.

\bibitem{xu2023sok}
Y.~Xu, X.~Han, G.~Deng, J.~Li, Y.~Liu, and T.~Zhang, ``Sok: Rethinking sensor
  spoofing attacks against robotic vehicles from a systematic view,'' in
  \emph{IEEE European Symposium on Security and Privacy}.\hskip 1em plus 0.5em
  minus 0.4em\relax IEEE, 2023, pp. 1082--1100.

\bibitem{JMLR:v13:gretton12a}
A.~Gretton, K.~M. Borgwardt, M.~J. Rasch, B.~Sch{{\"o}}lkopf, and A.~Smola, ``A
  kernel two-sample test,'' \emph{Journal of Machine Learning Research},
  vol.~13, no.~25, pp. 723--773, 2012.

\bibitem{zhou2018voxelnet}
Y.~Zhou and O.~Tuzel, ``Voxelnet: End-to-end learning for point cloud based 3d
  object detection,'' in \emph{IEEE Conference on Computer Vision and Pattern
  Recognition}, 2018, pp. 4490--4499.

\bibitem{chiang2020performance}
K.-W. Chiang, G.-J. Tsai, H.-J. Chu, and N.~El-Sheimy, ``Performance
  enhancement of ins/gnss/refreshed-slam integration for acceptable lane-level
  navigation accuracy,'' \emph{IEEE Transactions on Vehicular Technology},
  vol.~69, no.~3, pp. 2463--2476, 2020.

\bibitem{43405}
I.~Goodfellow, J.~Shlens, and C.~Szegedy, ``Explaining and harnessing
  adversarial examples,'' in \emph{International Conference on Learning
  Representations}, 2015.

\bibitem{kurakin2018adversarial}
A.~Kurakin, I.~J. Goodfellow, and S.~Bengio, ``Adversarial examples in the
  physical world,'' in \emph{Artificial intelligence safety and
  security}.\hskip 1em plus 0.5em minus 0.4em\relax Chapman and Hall/CRC, 2018,
  pp. 99--112.

\bibitem{madry2018towards}
A.~Madry, A.~Makelov, L.~Schmidt, D.~Tsipras, and A.~Vladu, ``Towards deep
  learning models resistant to adversarial attacks,'' in \emph{International
  Conference on Learning Representations}, 2018.

\bibitem{chen2019f}
Q.~Chen, X.~Ma, S.~Tang, J.~Guo, Q.~Yang, and S.~Fu, ``F-cooper: Feature based
  cooperative perception for autonomous vehicle edge computing system using 3d
  point clouds,'' in \emph{ACM/IEEE Symposium on Edge Computing}, 2019, pp.
  88--100.

\bibitem{lang2019pointpillars}
A.~H. Lang, S.~Vora, H.~Caesar, L.~Zhou, J.~Yang, and O.~Beijbom,
  ``Pointpillars: Fast encoders for object detection from point clouds,'' in
  \emph{IEEE Conference on Computer Vision and Pattern Recognition}, 2019, pp.
  12\,697--12\,705.

\bibitem{loshchilov2017decoupled}
I.~Loshchilov and F.~Hutter, ``Decoupled weight decay regularization,'' in
  \emph{International Conference on Learning Representations.}, 2017.

\end{thebibliography}

\end{document}